\documentclass[]{article}

\usepackage{graphicx}
\usepackage{amsmath}

\begin{document}

\title{Reflective visualization and verbalization of unconscious preference}

\author{Yoshiharu Maeno\footnote{Corresponding author. Graduate School of Systems Management, Tsukuba University, Otsuka 3-29-1, Bunkyo-ku, Tokyo 112-0012, Japan. E-mail: maeno.yoshiharu@nifty.com. Yoshiharu Maeno received the B.S. and M.S. degrees in physics from the University of Tokyo, Tokyo, Japan, and Ph.D. degree in business science from Tsukuba University. He is a management consultant and a founder of Social Design Group. He is a principal researcher at NEC Corporation. He seeks to invent a complexity-science-based methodology to design social systems and analyze social phenomena, aiming at solving diverse ranges of social and human problems. His research interests lie in understanding the stochastic temporal behaviors of non-linearly interacting components, the diffusion and reaction phenomena in innovative social organizations, and the latent dynamic structures behind social interactions. He is a member of the American Physical Society, the International Network of Social Network Analysis, and the Institute of Electrical and Electronic Engineers (Computer, Computational Intelligence, and Systems Man \& Cybernetics Societies).} and Yukio Ohsawa\footnote{School of Engineering, University of Tokyo, Hongo 7-3-1, Bunkyo-ku, Tokyo 113-8563, Japan. E-mail: ohsawa@sys.t.u-tokyo.ac.jp. Yukio Ohsawa received the Ph.D. degree in communication and information engineering from the University of Tokyo, Tokyo, Japan. He was with the Graduate School of Business Sciences, Tsukuba University, Tokyo. In 2005, he joined the School of Engineering, University of Tokyo, where he is currently an Associate Professor. He initiated the research area of chance discovery as well as a series of international meetings (conference sessions and workshops) on chance discovery, e.g., the fall symposium of the American Association of Artificial Intelligence (2001). He co-edited books on chance discovery published by Springer-Verlag and Advanced Knowledge International, and also special issues of journals such as New Generation Computing. Since 2003, his activity as Director of the Chance Discovery Consortium Japan has linked researchers in cognitive science, information sciences, and business sciences, and business people to chance discovery. It also led to the introduction of these techniques to researchers in Japan, the U.S., the U.K., China, Taiwan, R.O.C., etc.}}

\maketitle

\begin{abstract}
A new method is presented, that can help a person become aware of his or her unconscious preferences, and convey them to others in the form of verbal explanation. The method combines the concepts of reflection, visualization, and verbalization. The method was tested in an experiment where the unconscious preferences of the subjects for various artworks were investigated. In the experiment, two lessons were learned. The first is that it helps the subjects become aware of their unconscious preferences to verbalize weak preferences as compared with strong preferences through discussion over preference diagrams. The second is that it is effective to introduce an adjustable factor into visualization to adapt to the differences in the subjects and to foster their mutual understanding.
\end{abstract}

\section{Introduction}
\label{Introduction}

Every person has unique preferences that form the basis of the person's decision-making and consequent behaviors. However, when requested to describe the details of a preference that results in a behavior of particular interest to others, the person often fails to verbally explain. The person is not aware of all aspects of the preference. By understanding such unconscious influences, the person's private and social lifestyles can be re-designed \cite{Mae07a}. If they know the consumers' unconscious preferences, vendors can turn to new concepts, unfamiliar products, and emerging services to reach the consumers \cite{Zal03}. It is therefore important to develop methods to help a person become aware of his or her unconscious preferences and convey them to others in the form of verbal explanation. 

Various methods have been proposed in marketing to help us understand and foresee individual consumers' behavior. Conjoint analysis is used to discover the optimal combination of factors that customers would prefer \cite{Gre81}. The hierarchical Bayes model is employed to treat the different personalities of individual consumers. This model is applied to the investigation of the non-primary aspects of individual consumers' demand \cite{Aro98}. A number of studies on latent class models (or latent trait models) address latent variables and their statistical testing \cite{Che04}, \cite{Zha04}. These models are applied to test hypotheses on the factors in the consumers' preferences. They cannot, however, be used to discover unknown factors hidden in unconscious preferences.

Reflection in cognitive science \cite{Sch06} and computer-mediated communication \cite{Thu04} are theoretical guides to approach unconscious preferences. The abilities to recognize and understand oneself, discover something unexpected, and create something new are founded on constructive perception \cite{Suw02}, \cite{Suw03}. Constructive perception is the ability to perceive the visual characteristics of elements, the relationships between the elements, and the empty relevant space between the elements in diagrams, sketches, or drawings \cite{Lar87}. Visualization and verbalization play important roles in becoming aware of a person's present perception, and in changing it. Perception is a process used to interpret sensory signals from the outside. For example, drawing for reflection, which records a designer's daily intermediate outcomes, was proven to be an effective tool in a university education program of creativity \cite{Ish02}. A practical tool based on the theory of constructive perception is needed, like the drawings for designers, which can help a person's reflective visualization and verbalization of that person's unconscious preferences.

Methods of discovery in other fields can be applied to the discovery of unconscious preference. For example, in social network analysis, a heuristic method was developed to solve a node discovery problem. Its aim is to discover an unknown relevant person hidden in a criminal organization \cite{Kre02}. The person is not found in the records of observed actions but plays a relevant role in organizational communication and decision-making \cite{Mae08}. The method is implemented as an iterative process where the discrepancy between prior understanding and observation is indicated in the form of a social network diagram \cite{Mae07b}. Other methods of discovery are link prediction \cite{Cla08}, discovery of hierarchies \cite{Sal07}, cluster discovery \cite{Pal05}, and exploration of unknown structures \cite{New07}. Incorporating insights from the cognitive sciences would help to develop such a discovery process for treating unconscious preferences.

The objective of this paper is to develop a method that can help a person become aware of his or her unconscious preferences. The method combines the concepts of reflection, visualization (with an algorithm to draw the subject's stated preference in a diagram), and verbalization (through group discussion). The method is described in section \ref{method} (with the preference diagram in \ref{Preference}, the visualization algorithm in \ref{visualization}, and the reflection process in \ref{reflection}). The experiment of testing the reflection process to investigate the unconscious preferences of subjects with artwork and the lessons learned from it are presented in section \ref{Experiment}.

\section{Method}
\label{method}

\subsection{Preference diagram}
\label{Preference}

\begin{figure}
\begin{center}
\includegraphics[scale=0.48,angle=-90]{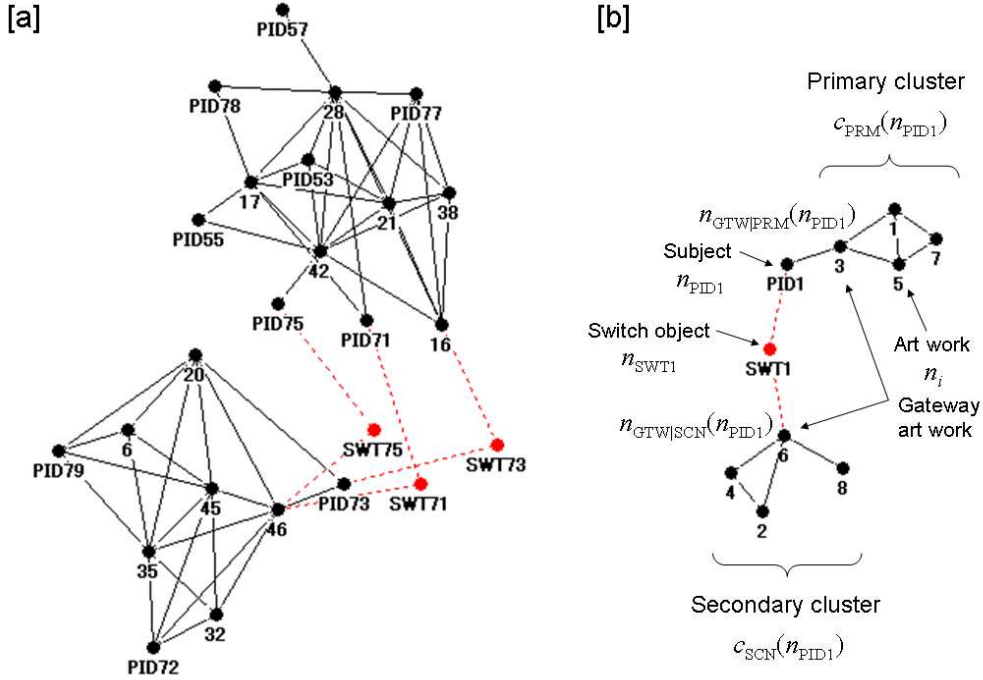}
\end{center}
\caption{[a] Example of a preference diagram, and [b] characteristic objects (the primary cluster $c_{{\rm PRM}}(n_{{\rm PID}1})$, the secondary cluster $c_{{\rm SCN}}(n_{{\rm PID}1})$, and the gateway art works $n_{{\rm GTW|PRM}}(n_{{\rm PID}1})$, $n_{{\rm GTW|SCN}}(n_{{\rm PID}1})$), and a switch object $n_{{\rm SWT}1}$ for a subject person $n_{{\rm PID}1}$.}
\label{RVVUP4}
\end{figure}

A preference diagram is a practical tool that can be used to promote the reflective visualization and verbalization of unconscious preferences. It also aids in efficient communication and mutual understanding in group discussion. The preference diagram is a kind of undirected graph consisting of nodes and undirected links drawn on a plane. A node represents either the $i$-th subject $n_{{\rm PID}i}$ or the $j$-th artwork $n_{j}$. A link represents either the resemblance relationship between two artworks or the preference relationship of a subject to an artwork. The topology defined by the presence of links is relevant. The position of the nodes and the distance between the nodes on the diagram, however, are not relevant. Figure \ref{RVVUP4} [a] shows an example.

The preference diagram is designed so that the cluster structures are clearly visible. For an individual subject $n_{{\rm PID}i}$, four characteristic objects are selected as attributes describing the subject's preferences. The objects are the primary cluster $c_{{\rm PRM}}(n_{{\rm PID}i})$, the gateway artwork in the primary cluster $n_{{\rm GTW|PRM}}(n_{{\rm PID}i})$, the secondary cluster $c_{{\rm SCN}}(n_{{\rm PID}i})$, and the gateway artwork in the secondary cluster $n_{{\rm GTW|SCN}}(n_{{\rm PID}i})$. The primary cluster represents the cluster whose artworks are preferred by a subject most strongly. The secondary cluster is the cluster whose artworks are preferred most weakly. The gateway artwork indicates a subject's preferable entrance point to the cluster. Note that the primary cluster for a subject can be the secondary cluster for another subject. The labels for the characteristic objects are not drawn on the diagram. There may be multiple gateway artworks.

The preference diagram includes another interesting object. A switch object $n_{{\rm SWT}i}$ for a subject $n_{{\rm PID}i}$ is inserted into the space between the primary and secondary clusters. The switch is assumed to change the mode of the subject's preference. The switch tends to point to the primary cluster most of the time, although it sometimes points to the secondary cluster. The switching occurs either from internal caprice or from an external stimulus such as atmosphere, influence from friends, or social interaction. The switch indicates the presence of an unknown factor that prompts the subjects to turn their interests toward something unfamiliar. This is the clue to unconscious preferences. Figure \ref{RVVUP4} [b] shows an example of the four characteristic objects and the switch object for a subject person $n_{{\rm PID}1}$. 

\subsection{Visualization algorithm}
\label{visualization}

A visualization algorithm to generate a preference diagram from the information about a subject's stated preference is presented. The information can be obtained from the answers to questionnaires. It is a set of data \mbox{\boldmath{$D$}} given by Equation (\ref{answers1}). 

\begin{equation}
\mbox{\boldmath{$D$}} = \{d_{l} \} \ (0 \leq l \leq |d|-1).	
\label{answers1}
\end{equation}

An individual datum $d_{l}$ represents an answer from a subject. The number of data is $|d|$. The answer is formatted as in Equation (\ref{answers2}). 

\begin{equation}
d_{l}  = n_{{\rm PID}i} \oplus \{ n_{j} \}.	
\label{answers2}
\end{equation}

It is a combination of the identifier of the subject ($n_{{\rm PID}i}$) and a set of identifiers for any number of artworks that the subject selected as preferable ($\{n_{j}\}$). We denote such a combination by $\oplus$. In general, a single subject may answer multiple times at different occasions, resulting in multiple data. The subject is, however, allowed to answer only once for the purposes of this paper. The number of data $|d|$ is the same as the number of subjects $|s|$. We order the data $d_{l}$ in \mbox{\boldmath{$D$}} so that the relation $l=i$ can hold.

At first, the artworks are grouped into clusters $c_{k}$. The number of clusters $|c|$ is given. An individual cluster includes the artworks which resemble the stated preference. This is interpreted as the granularity of the preference diagram. As the granularity becomes finer, the number of clusters $|c|$ increases, but the number of artworks in a cluster decreases. A clustering algorithm for discrete objects is applied for given $|c|$. The k-medoids algorithm is a simple example \cite{Has01}. A medoid is an object that is the closest to the center of gravity in a cluster. Its principle is similar to that of the k-means algorithm \cite{Dud00} for continuous numerical variables where the center of gravity is updated repeatedly according to the expectation-maximization method \cite{Dem77}. The degree of resemblance for every pair of artworks $n_{i}$ and $n_{j}$ is given by the Jaccard coefficient $J(n_{i},n_{j})$ defined by Equation (\ref{Jaccard}). 

\begin{equation}
J(n_{i},n_{j}) \equiv \frac{F(n_{i} \cap n_{j})}{F(n_{i} \cup n_{j})}.
\label{Jaccard}
\end{equation}

The Jaccard coefficient is a measure of co-occurrence that is employed in link discovery problems, text document analysis, or WWW structure mining \cite{Lib07}. The function $F(n_{j})$ is the occurrence frequency at which the artwork $n_{j}$ appears in $\mbox{\boldmath{$D$}}$. In this paper, it is the same as the number of subjects who selected the artwork $n_{j}$ as preferable. Equation (\ref{Jaccard}) can be converted into Equation (\ref{Jaccard2}) with a Boolean function $B(s)$ in Equation (\ref{Boo}). $B(s)$ determines whether the proposition $s$ is true or false.

\begin{equation}
J(n_{i},n_{j}) = \frac{\sum_{l=0}^{|d|-1} B( n_{i} \in d_{k} \wedge n_{j} \in d_{k}) }{ \sum_{l=0}^{|d|-1} B( n_{i} \in d_{k} \vee n_{j} \in d_{k} ) }.
\label{Jaccard2}
\end{equation}

\begin{equation}
B(s) = \left \{ \begin{array}{ll}
          1 & \mbox{\ \ if $s$ is TRUE} \\
          0 & \mbox{\ \ otherwise}
        \end{array}
    \right ..
\label{Boo}
\end{equation}

Initially, the artworks are grouped into clusters at random. The medoids $n_{{\rm MED}}(c_{k})$ in the cluster $c_{k}$ are calculated by Equation (\ref{medoid1}).

\begin{equation}
n_{{\rm MED}}(c_{k}) = {\rm arg} \ \underset{n_{j} \in c_{k}}{{\rm max}} \ M(c_{k},n_{j}) \ (0 \leq k \leq |c|-1).
\label{medoid1}
\end{equation}

The operator ${\rm arg}$ in Equation (\ref{medoid1}) means that the medoid is the node $n_{j}$ belonging to $c_{k}$, which maximizes $M(c_{k},n_{j})$. The quantity $M(c_{k},n_{j})$ in Equation (\ref{medoid1}) represents the total degree of resemblance of one artwork $n_{j}$ to the other artworks in the cluster $c_{k}$. It is defined by Equation (\ref{medoid2}). 

\begin{equation}
M(c_{k},n_{j}) = \sum_{ {\scriptsize \begin{array}{c} n_{l} \in c_{k} \\ n_{l} \neq n_{j} \end{array} } } J(n_{l},n_{j}).
\label{medoid2}
\end{equation}

After the medoids are determined, the artworks are regrouped into clusters. The cluster $c(n_{j})$ to which an artwork $n_{j}$ belongs is calculated by Equation (\ref{medoid3}).

\begin{equation}
c(n_{j}) = {\rm arg} \ \underset{c_{k}}{{\rm max}} \ J(n_{{\rm MED}}(c_{k}),n_{j}).
\label{medoid3}
\end{equation}

The calculation of the medoids in Equation (\ref{medoid1}) and the clusters to which the artworks belong in Equation (\ref{medoid3}) is repeated until they converge. After that, links are drawn between the nodes $n_{i}$ and $n_{j}$ belonging to a cluster, if $J(n_{i},n_{j})>0$. These links and nodes form $|c|$ disjoint clusters. Any artwork that is not selected by any subjects becomes an isolated node. 

After the clustering is complete, the primary cluster $c_{{\rm PRM}}(n_{{\rm PID}i})$, the secondary cluster $c_{{\rm SCN}}(n_{{\rm PID}i})$, and the gateway artworks for each individual subject $n_{{\rm PID}i}$ are calculated. The primary cluster $c_{{\rm PRM}}(n_{{\rm PID}i})$ is calculated by Equation (\ref{Pri1}).

\begin{equation}
c_{{\rm PRM}}(n_{{\rm PID}i}) = {\rm arg} \ \underset{c_{k}}{{\rm max}} \ \underset{n_{j} \in c_{k}}{{\rm max}} W(n_{{\rm PID}i},n_{j}).
\label{Pri1}
\end{equation}

The operator ${\rm arg}$ in Equation (\ref{Pri1}) means the following. The maximal value of $W(n_{{\rm PID}i},n_{j})$ is searched for among all the artworks $n_{j}$ belonging to the cluster $c_{k}$. The primary cluster $c_{{\rm PRM}}(n_{{\rm PID}i})$ is the cluster that gives the maximal value of ${\rm max} \ W(n_{{\rm PID}i},n_{j})$ among the clusters $c_{k}$. $W(n_{{\rm PID}i},n_{k})$ in Equation (\ref{Pri1}) represents the strength of the preference of the subject $n_{{\rm PID}i}$ to the artwork $n_{k}$. It is defined by Equation (\ref{Wei}). 

\begin{equation}
W(n_{{\rm PID}i},n_{j}) = \frac{\sum_{l=0}^{|d|-1} B(n_{j} \in d_{l} \wedge n_{{\rm PID}i} \in d_{l})}{\sum_{l=0}^{|d|-1} B(n_{j} \in d_{l})}.
\label{Wei}
\end{equation}

The subject answers once, and the relation $l=i$ holds in Equation (\ref{answers2}). Equation (\ref{Wei}) becomes simpler because $B(n_{{\rm PID}i} \in d_{l})=\delta_{li}$.

\begin{equation}
W(n_{{\rm PID}i},n_{j}) = \frac{B(n_{j} \in d_{i})}{\sum_{l=0}^{|d|-1} B(n_{j} \in d_{l})}.
\label{Wei2}
\end{equation}

The gateway artwork in the primary cluster $n_{{\rm GTW|PRM}}(n_{{\rm PID}i})$ is calculated by Equation (\ref{Pri2}).

\begin{equation}
n_{{\rm GTW|PRM}}(n_{{\rm PID}i}) = {\rm arg} \ \underset{n_{j} \in c_{{\rm PRM}}(n_{{\rm PID}i})}{{\rm max}} W(n_{{\rm PID}i},n_{j}).
\label{Pri2}
\end{equation}

The operator ${\rm arg}$ means that $n_{{\rm GTW|PRM}}({\rm PID}_{i})$ is the artwork that gives the maximal value of $W(n_{{\rm PID}i},n_{k})$ among $n_{k}$ belonging to the primary cluster $c_{{\rm PRM}}(n_{{\rm PID}i})$. There may be multiple gateway artworks. Links are drawn between the subject and the gateway artworks in the primary cluster.

The secondary cluster $c_{{\rm SCN}}(n_{{\rm PID}i})$ is calculated by Equation (\ref{Sec3}).

\begin{equation}
c_{{\rm SCN}}(n_{{\rm PID}i}) = {\rm arg} \ \underset{c_{k}}{{\rm min}} \ \ \underset{n_{j} \in c_{k}}{{\rm max}} W(n_{{\rm PID}i},n_{j}).
\label{Sec3}
\end{equation}

It is the cluster whose artworks are preferred by the subject most weakly. Alternatively, the secondary cluster can be the cluster whose artworks are preferred by the subject most strongly after the primary cluster. It is calculated by Equation (\ref{Sec1}) instead of Equation (\ref{Sec3}). 

\begin{equation}
c_{{\rm SCN}}(n_{{\rm PID}i}) = {\rm arg} \ \underset{c_{k} \neq c_{{\rm PRM}}(n_{{\rm PID}i})}{{\rm max}} \ \ \underset{n_{j} \in c_{k}}{{\rm max}} W(n_{{\rm PID}i},n_{j}).
\label{Sec1}
\end{equation}

The gateway artwork in the secondary cluster $n_{{\rm GTW|SCN}}(n_{{\rm PID}i})$ is calculated by Equation (\ref{Sec2}). All characteristic objects are determined here. 

\begin{equation}
n_{{\rm GTW|SND}}(n_{{\rm PID}i}) = {\rm arg} \ \underset{n_{j} \in c_{{\rm SND}}(n_{{\rm PID}i})}{{\rm max}} W(n_{{\rm PID}i},n_{j}).
\label{Sec2}
\end{equation}

Finally, links are drawn between the disjoint clusters so that the switch object $n_{{\rm SWT}i}$ can connect the subject $n_{{\rm PID}i}$ and the gateway artwork in the secondary cluster $n_{{\rm GTW|SND}}(n_{{\rm PID}i})$, as in Figure \ref{RVVUP4} [b]. The preference diagram uses the spring model \cite{Fru91} as a graph-drawing method. The spring model converts the strength of the relationship across the link between two nodes into Hooke's constant of the spring, which is placed between the nodes imaginarily, and calculates the equilibrium position of the nodes.

\subsection{Reflection process}
\label{reflection}

\begin{figure}
\begin{center}
\includegraphics[scale=0.48,angle=-90]{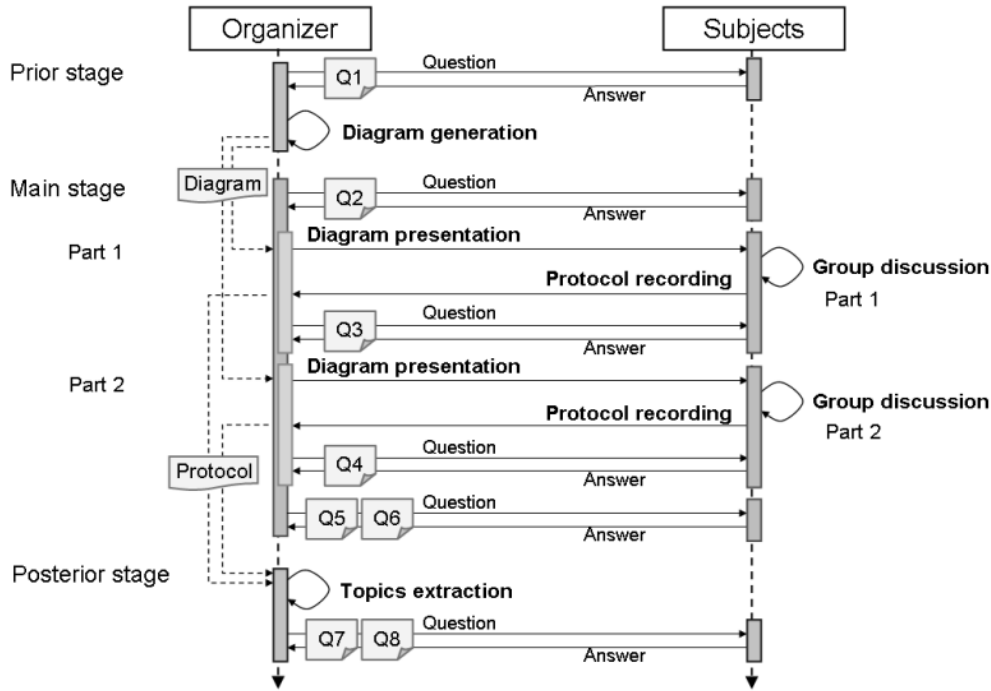}
\end{center}
\caption{Sequence in the reflection process. The notations follow the UML (unified modeling language) specifications. The time goes by from the top to the bottom. The process consists of the 3 stages. The organizer asks the subjects a questionnaire (consisting of 8 questions Q1 to Q8). The organizer generates the preference diagrams from the answers to the question Q1 after the prior stage. The organizer extracts the topics, in which the subjects are interested in the group discussions, from the recorded protocols after the main stage.}
\label{RVVUP3}
\end{figure}

The reflection process uses the preference diagrams to prompt the subjects' reflective visualization and verbalization. Group discussion is incorporated into the process as a means to understand the frame of the subjects' own perception. The diagram is also meant to help create efficient communication and mutual understanding during group discussion. The sequence in the designed reflection process is shown by Figure \ref{RVVUP3}. The notations follow the UML (unified modeling language) \cite{UML} specifications. It defines a graphical language for visualizing, specifying, constructing, and documenting the artifacts of distributed object systems. The time goes by from top to bottom. The sequence consists of the prior, main, and posterior stages. An organizer coordinates the process. The organizer uses a a questionnaire to ask the subjects eight questions (Q1 to Q8). Questions Q1 to Q4 are essential for the subjects to become aware of their unconscious preference. The other questions (Q5 to Q8) are for the purpose of evaluating the designed reflection process in the experiment.

The content of questions Q1 to Q8 is listed in Table \ref{table1}. The organizer generates the preference diagrams from the answers to question Q1 after the prior stage. Group discussions are carried out in two sub-stages (part 1 and part 2). This is for the purpose of evaluating the experiment. The number of sub-stages can be one or arbitrary. In the group discussions, the organizer asks the subjects to discuss the preferences of themselves or the others freely. Drawing any conclusion is not requested. Questions Q2 to Q4 are the central drivers in this process, which promotes reflection. The organizer extracts the topics in which the subjects express interest during the group discussions from the recorded protocols after the main stage. The protocols are the verbal reports from the subjects. The topics are used in questions Q7 and Q8 in the posterior stage.

\begin{table}
\begin{center}
\begin{tabular}{|c|l|}
\hline
\# & Question \\
\hline
Q1 & Which art works do you prefer? \\
\hline
Q2 & What is your preference on the art works? \\
\hline
Q3 & What did you become aware of on your preference? \\
\hline
Q4 & What did you become aware of on your preference? \\
\hline
Q5 & Were the preference diagrams used in the part I discussion useful? \\
\hline
Q6 & Were the preference diagrams used in the part II discussion useful? \\
\hline
Q7 & Did the individual topic appearing in the discussions help you verify \\
  & the understanding of your preference? \\
\hline
Q8 & Did the individual topic appearing in the discussions help you become \\
  & aware of your preference? \\
\hline
\end{tabular}
\end{center}
\caption{The 8 questions (Q1 to Q8) requested to the subjects in Figure \ref{RVVUP3}.}
\label{table1}
\end{table}

\section{Experiment}
\label{Experiment}

\subsection{Visualization}

\begin{figure}
\begin{center}
\includegraphics[scale=0.5,angle=-90]{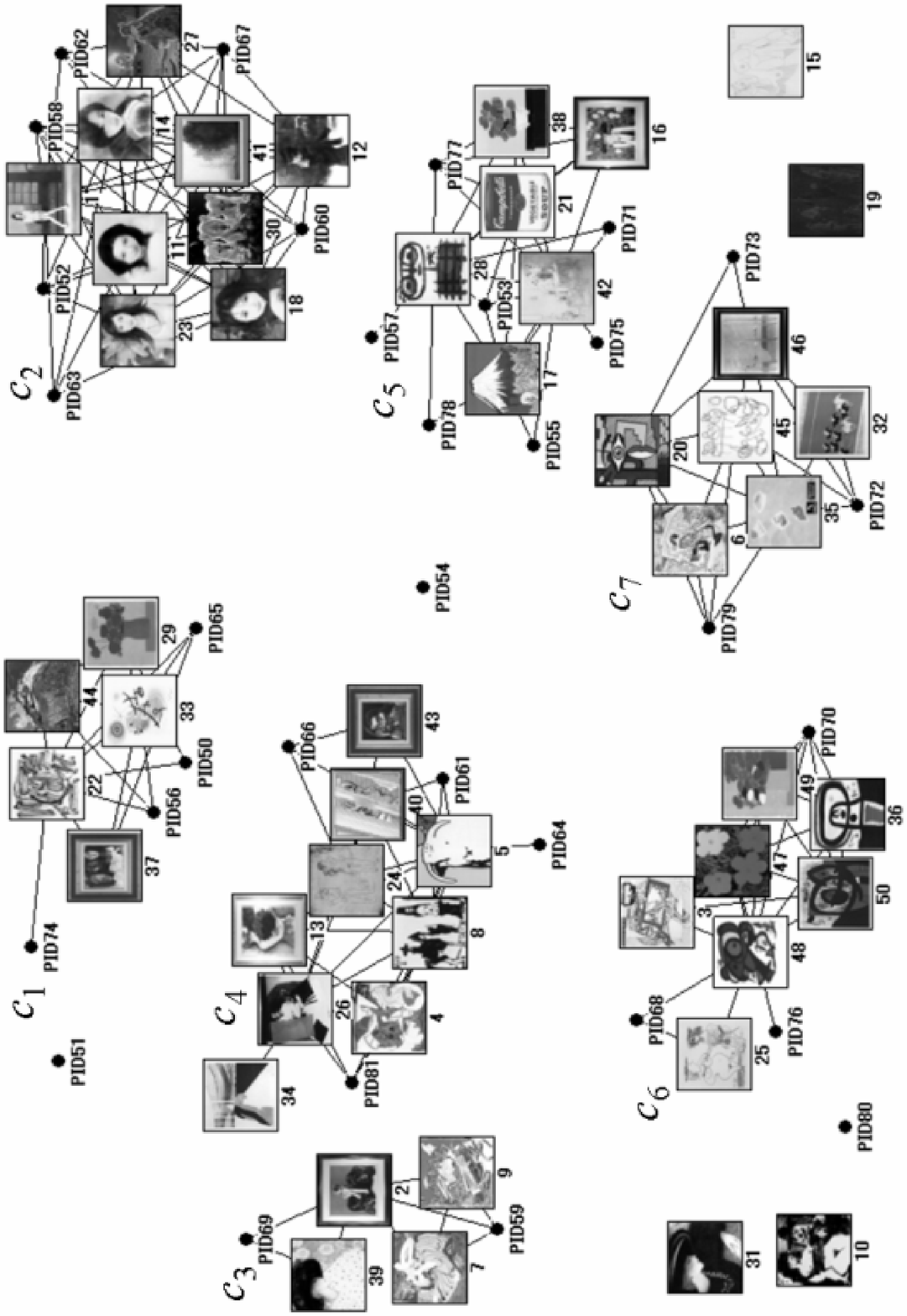}
\end{center}
\caption{Example of the preference diagrams which are used in the part 1 group discussion. The diagram indicates the cluster characteristics among the art works, and the subjects (such as $n_{{\rm PID}50}$). The number of clusters is $|c|=7$. }
\label{RVVUP1}
\end{figure}

The experiment was carried out according to the reflection process described in \ref{reflection}. Fifty artworks (classical portraits, landscapes, abstract paintings, modern pop art) are used in Q1 in Figure \ref{RVVUP3}. Thirty-two subjects participated in the prior stage. The coordinator generated preference diagrams as presented in \ref{visualization}. The main stage was carried out three separate times, with four, two, and five subjects. It took sixty to ninety minutes to finish the main stage.

The four diagrams that include the cluster structures were presented in the part 1 group discussion. Finer granularity diagrams (the number of clusters $|c|$=3, 5) and courser granularity diagrams ($|c|$=7, 8) were presented at the same time. The subjects could recognize the primary clusters, compare the details of the diagrams, and discuss them freely. The diagram of $|c|=7$ is shown by Figure \ref{RVVUP1}. Images of the artworks are attached to the corresponding nodes. 

\begin{figure}
\begin{center}
\includegraphics[scale=0.5,angle=-90]{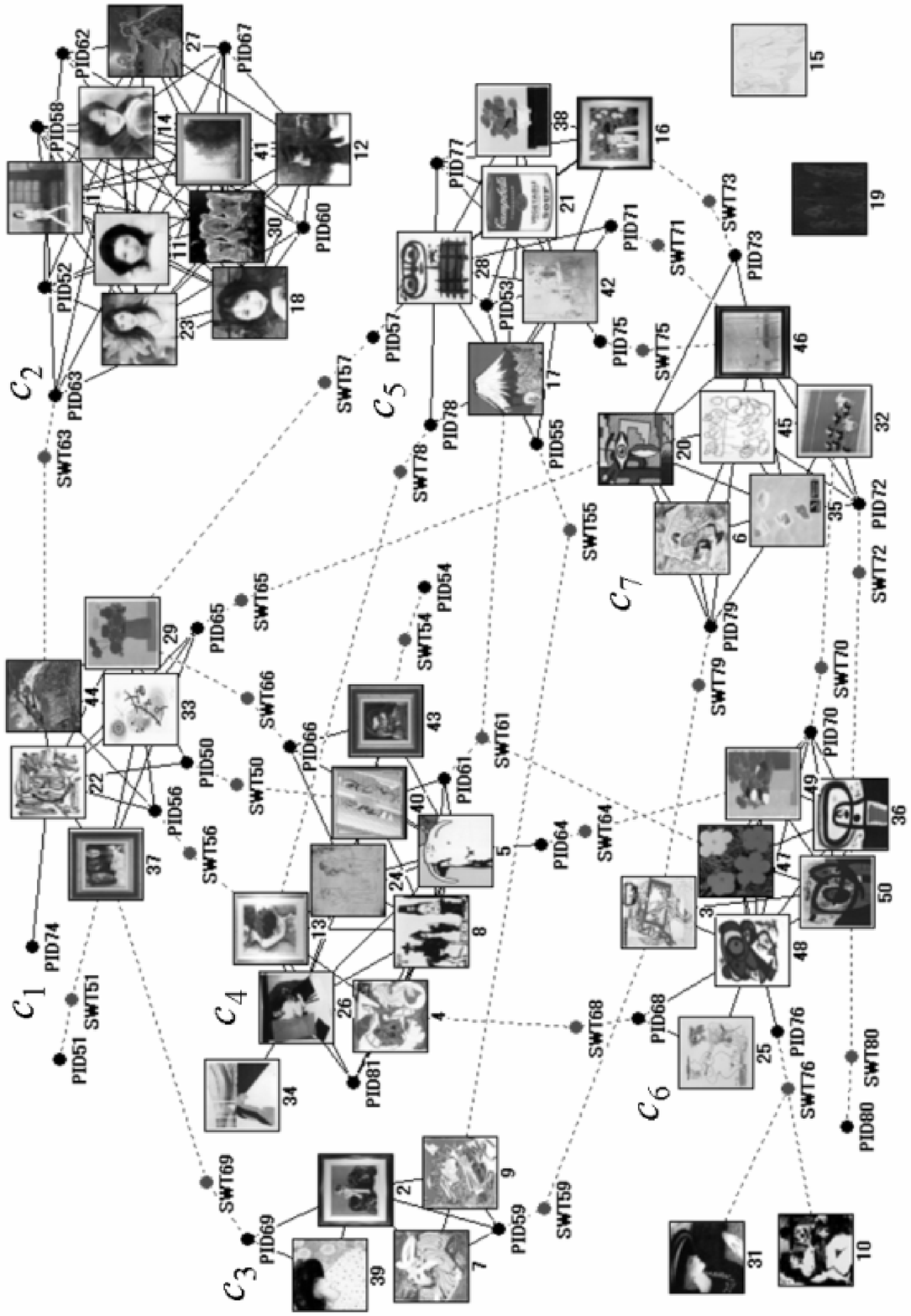}
\end{center}
\caption{Example of the preference diagrams which are used in the part 2 of the group discussions. The diagram indicates the relationships among the art works, the subjects, and the switch objects (such as $n_{{\rm SWT}50}$ for $n_{{\rm PID}50}$) from the subjects in the primary clusters toward the secondary clusters. The number of clusters is $|c|=7$.}
\label{RVVUP2}
\end{figure}

The four diagrams that include the switch objects were presented in the part 2 group discussion. Finer and coarser granularity diagrams ($|c|$=3, 5, 7, and 8) were presented at the same time as in the part 1 group discussion. The diagram of $|c|=7$ is shown by Figure \ref{RVVUP2}. The subjects could recognize the switch objects, compare the strong and weak preferences, and obtain a clue to their unconscious preferences by interpreting and explaining the diagrams in the group discussion.

The answers to questions Q2 to Q8 were used in three analyses. Questions Q5 and Q6 are for the first analysis: the evaluation of the reflection process. Questions Q7 and Q8 are for the second analysis: the analysis of the characteristics of the topics that helped the subjects become aware of their unconscious preferences. Questions Q2 to Q4 are for the third analysis: the analysis of the characteristics of the unconscious preferences of which the subjects became conscious. The second and third analyses are demonstrated in \ref{verbalization}.

Here, the first analysis is demonstrated. A summary of the answers to questions Q5 and Q6 are shown in Table \ref{table2}. Every subject agreed that the preference diagrams were useful in the part 1 group discussion. Most subjects agreed that they were useful in the part 2 group discussion. The answers indicate that the visualization tool, like the preference diagram, is useful in general. The reason some did not realize their unconscious preferences is analyzed below with the third analysis.

\begin{table}
\begin{center}
\begin{tabular}{|l|c|c|}
\hline
\# & Number of YES & Ratio \\
\hline
Q5 & 11 & 100\% \\
\hline
Q6 & 9 & 82 \% \\
\hline
\end{tabular}
\end{center}
\caption{Summary of the answers to the questions Q5 and Q6.}
\label{table2}
\end{table}

\subsection{Verbalization}
\label{verbalization}

In the second analysis, the group discussions were recorded as protocols. The coordinator extracted the relevant topics of the discussions from the recorded protocols. The relevant topics are those on which the subjects discussed for more than five minutes. The nine extracted topics are listed in Table \ref{table3}. Among them, five topics appeared in the part 1 discussion. They are related to the interpretation of the clusters, individual subjects' personalities, and the change arising from the granularity. The others appeared in the part 2 discussion. The subjects were interested in the interpretation of the switch objects and the secondary clusters. Discussion on the discrepancy among individual subjects' expectations and interpretations and the diagrams were hot throughout the discussions.

\begin{table}
\begin{center}
\begin{tabular}{|c|c|l|}
\hline
 & \# & Topic \\
\hline
 & T1 & How do the expression and the meaning of the art works \\
 &    & belonging to the clusters resemble? \\
\cline{2-3}
 & T2 & How does the personality of the subjects belonging \\
 &    & to the clusters resemble? \\
\cline{2-3}
Part 1 & T3 & Do you like or dislike the clusters? \\
\cline{2-3}
 & T4 & Are the clusters different from the art works which you \\
 &    & feel are similar? \\
\cline{2-3}
 & T5 & How do the clusters look differently when the granularity \\
 &    & is changed? \\
\hline
 & T6 & What do the clusters, which look unchanged when the \\
 &    & granularity is changed, mean? \\
\cline{2-3}
 & T7 & What do the switch objects represent? \\
\cline{2-3}
Part 2 & T8 & Do you like or dislike the secondary cluster to which \\
 &    & the switch object has link? \\
\cline{2-3}
 & T9 & How do the clusters look differently when the granularity \\
 &    & is changed? \\
\hline
\end{tabular}
\end{center}
\caption{The 9 discussion topics extracted from the protocol analysis.} 
\label{table3}
\end{table}

The answers to questions Q7 and Q8 are summarized in Table \ref{table4}. The top three topics as ranked in Q7 are topics T3, T2, and T1. The topics that were extracted from the part 1 discussion were useful in helping the subjects verify the understanding of their preferences. Topics T7, T8, and T9, which were extracted from the part 2 discussion, were not selected at all. The subjects could be convinced that their understanding agreed with the others' if the clustering structure in the diagrams could be interpreted as easily as they expected.

The top three topics ranked in question Q8 were the topics T8, T9, and T7. The topics extracted from the part 2 discussion helped the subjects become aware of their preferences. The ranking of the topics from Q2 and Q1 becomes low while the ranking of the topics from Q3 is high both in Q7 and Q8. The subjects could become aware of their unconscious preferences by comparing the weak preference (the secondary cluster) with the strong preference (the primary cluster), and by attempting to verbalize the origin and background (the switch object) of the weak preference. Visualizing and verbalizing the weak preference in contrast to the strong preference contributes to becoming aware of the unconscious preference. The preference diagrams used in the part 2 discussion are more effective for this purpose. These are the first lessons learned in the experiments.

\begin{table}
\begin{center}
\begin{tabular}{|c|c|c|c|c|}
\hline
 & \multicolumn{4}{c|}{Question} \\
\cline{2-5}
Topics & \multicolumn{2}{c|}{Q7} & \multicolumn{2}{c|}{Q8} \\
\cline{2-5}
 & Number of YES & Ranking & Number of YES & Ranking \\
\hline
T1 & 6 & 2 & 3 & 5 \\
\hline
T2 & 5 & 3 & 2 & 8 \\
\hline
T3 & 7 & 1 & 5 & 2 \\
\hline
T4 & 4 & 4 & 3 & 5 \\
\hline
T5 & 3 & 6 & 3 & 5 \\
\hline
T6 & 4 & 4 & 1 & 9 \\
\hline
T7 & 0 & (7) & 4 & 4 \\
\hline
T8 & 0 & (7) & 7 & 1 \\
\hline
T9 & 0 & (7) & 5 & 2 \\
\hline
\end{tabular}
\end{center}
\caption{Summary of the answers to the quetions Q7 and Q8.} 
\label{table4}
\end{table}

Next, the third analysis is demonstrated. Examples of the answers to questions Q2 to Q4 are listed in Table \ref{table5}. Subjects $n_{{\rm PID}59}$ and $n_{{\rm PID}72}$ did not answer YES to question Q6 in Table \ref{table2}. Subject $n_{{\rm PID}59}$ discovered that the understandable degree of abstractness and warm colors are relevant as a motif. According to the protocols, the subject talked about the primary clusters in the coarser granularity preference diagrams. The subject particularly observed the technical details of the artworks in the primary cluster. This seems to be the reason why the subject felt that the switch to the secondary clusters was not relevant. Subject $n_{{\rm PID}72}$ discovered that the simplicity the subject prefers means the calmness that heals the subject rather than the technical conciseness of the drawings. According to the protocols, the subject talked about the primary clusters in the finer granularity preference diagrams. Similarly to subject $n_{{\rm PID}59}$, attention was paid to observing the detailed expression that the artworks in the primary cluster conveys. This seems to be the reason why the switch is not relevant.

The other subjects answered YES to question Q6 in Table \ref{table2}. Subject $n_{{\rm PID}80}$ talked about the secondary clusters in the more coarse granularity preference diagrams, and became aware that the subject gets tired if the drawings include only living things or nature scenes. Subject $n_{{\rm PID}81}$ talked about the secondary clusters in the finer granularity preference diagrams, and became aware that the temporary feeling, or the influence from friends, is included in the factors that determine the weak preference. Subject $n_{{\rm PID}66}$ became aware that the flow of lines or brush strokes is relevant to the preference, rather than the composition, which the subject expected before the group discussion. Subject $n_{{\rm PID}71}$ talked about the artworks in the finer granularity preference diagrams and became aware that the subject has an intuitive sense of preference, which is contrary to the subject's prior understanding.

The content, which the subjects become aware of in the reflection process, ranges widely. It is not limited to the unconscious preference. It may be a deeper analysis of the expression in the artworks or the criteria to select the artworks which the subjects prefer strongly. The degree of the prior understanding of the subjects also ranges widely. The granularity of the preference diagram with which the individual subjects discover something depends on the subject. It is important to introduce an adjustable parameter in visualization such as granularity to adapt to the differences in the subjects. These are the second lessons learned in the experiments.

\begin{table}
\begin{center}
\begin{tabular}{|c|l|l|}
\hline
Subject & \# & Answers \\
\hline
 & Q2 & I don't like persons as a motif. \\
59 & Q3 & Understandable abstractness, warm colors, and landscape \\
 & & motifs are relevant to my preference. \\
 & Q4 & The same as the Q3. \\
\hline
 & Q2 & Composition, rather than color, governs the comforts I feel. \\
66 & Q3 & Motifs are not relevant to my preference. \\
 & & I don't like the art works like commercial posters. \\
 & Q4 & Flow of lines or paintbrushes determines the comforts I feel. \\
\hline
 & Q2 & I can't imagine the art works which I dislike. \\
 & Q3 & I noticed that I have preference than I expected. \\
71 & & I don't like the primary colors such as red or yellow. \\
 & Q4 & The diagrams agrees with my intuitive feeling more firmly \\
 & & as the granularity of the diagrams becomes finer. \\
\hline
 & Q2 & I like landscapes and the drawings which are not complicated. \\
72 & Q3 & I noticed that simplicity becomes the calmness which heals me. \\
 & Q4 & The same as the Q3. \\
\hline
 & Q2 & I like the art works which assert themselves in drawing nature, \\
 & & or make myself feel at ease. \\
80 & Q3 & I began to feel that the simplicity in the abstract painting \\
 & & is one of the assertions which attracts me. \\
 & Q4 & I noticed that I got tired if the drawing includes only \\
 &    & living things, or nature scenes. \\
\hline
 & Q2 & I like funny, understandable, impressive, or queer art works. \\
 & Q3 & I am surprised at some of the art works which belong \\
81 & & to my cluster, but fond of the others. \\
 & Q4 & The secondary cluster is related to my temporary feeling, \\
 & & or the influence from my friends. \\
\hline
\end{tabular}
\end{center}
\caption{Example of the answers to the questions Q2, Q3, and Q4.} 
\label{table5}
\end{table}

\section{Conclusion}
\label{Conclusion}

For this paper, we developed a method that helps people become aware of their unconscious preferences and convey them to others in the form of verbal explanation. The method combines the concepts of reflection, visualization (with an algorithm to draw the subjects' stated preferences in a diagram), and verbalization (through group discussion). The method was tested in an experiment where the unconscious preferences of the subjects for various artworks were investigated. 

In the experiment, two lessons were learned. The first is that it helps the subjects become aware of their unconscious preferences to verbalize weak preferences as compared with strong preferences through discussion over preference diagrams. The second is that it is effective to introduce an adjustable factor into visualization, such as the granularity of the preference diagram, to adapt to the differences in the subjects and to foster their mutual understanding. The lessons imply that the interpretation of the weak preferences, which emerged in the reflection process, is subject to the particular nature of the presented artworks and the subjects joining the discussion. The questionnaire may also affect the subjects. Which artwork is suitable for your room? Which artwork would you like to buy? These two questions may cause different responses from the subjects. Preferences may also be influenced by a change in the environment. The individual's unconscious tendency itself may change by the individual becoming aware of its presence. Such sensitivity is essential to modern consumer behavior. It is beneficial to focus on such case-by-case preferences rather than to make an effort to figure out the universal laws of human behavior.

Unconscious preference is one of the factors in stimulating brand switching in marketing or extending belonging groups in communication. We have taken the first step toward understanding individuals' unconscious tendencies in thinking and acting. Our method stimulates the existing unconscious tendency in order to prompt change. The method, which exploits the delicate nature of the unconscious preference, is essential in the future of marketing, education, communication, and many other applications.

\end{document}